# Optimizing colormaps with consideration for color vision deficiency to enable accurate interpretation of scientific data

Jamie R. Nuñez, Christopher R. Anderton, Ryan S. Renslow*

Earth and Biological Sciences Directorate, Pacific Northwest National Laboratory, Richland, Washington, United States of America

* ryan.renslow@pnnl.gov

## Abstract

Color vision deficiency (CVD) affects more than 4% of the population and leads to a different visual perception of colors. Though this has been known for decades, colormaps with many colors across the visual spectra are often used to represent data, leading to the potential for misinterpretation or difficulty with interpretation by someone with this deficiency. Until the creation of the module presented here, there were no colormaps mathematically optimized for CVD using modern color appearance models. While there have been some attempts to make aesthetically pleasing or subjectively tolerable colormaps for those with CVD, our goal was to make optimized colormaps for the most accurate perception of scientific data by as many viewers as possible. We developed a Python module, *cmaputil*, to create CVD-optimized colormaps, which imports colormaps and modifies them to be perceptually uniform in CVD-safe colorspace while linearizing and maximizing the brightness range. The module is made available to the science community to enable others to easily create their own CVD-optimized colormaps. Here, we present an example CVD-optimized colormap created with this module that is optimized for viewing by those without a CVD as well as those with red-green colorblindness. This colormap, cividis, enables nearly-identical visual-data interpretation to both groups, is perceptually uniform in hue and brightness, and increases in brightness linearly.

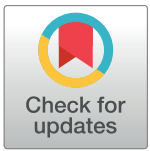

## Introduction

Presenting data that can be quickly interpreted and easily understood is essential in the scientific community. Often, a quick view of a study's results is the primary source of relaying information and gaining interest, making it critical for the author to consider how it will be interpreted [1–6]. Among other components, the colors chosen to relay data must be considered and applied carefully. Here, we focus on using appropriate colormaps, arrays of colors used in a pre-defined order, for representing i) digitally reconstructed scientific image data, ii) how someone with color vision deficiency (CVD) may be affected by this choice, and iii) how we can create an optimized colormap for those with CVD as well as those without.

**Funding:** This research was partially supported by the Genomic Science Program (GSP), Office of Biological and Environmental Research (OBER), the U.S. Department of Energy (DOE), and is a contribution of the Pacific Northwest National Laboratory (PNNL) Foundational Scientific Focus Area (SFA). A portion of this research was performed in the W. R. Wiley Environmental Molecular Sciences Laboratory (EMSL), a national scientific user facility sponsored by the Office of Biological and Environmental Research (BER) and located at PNNL. PNNL is a multi-program national laboratory operated by Battelle for the DOE under Contract DE-AC05-76RLO 1830.The funders had no role in study design, data collection and analysis, decision to publish, or preparation of the manuscript.

**Competing interests:** The authors have declared that no competing interests exist.

CVD reduces the ability to distinguish between certain colors and affects up to 8% of men and 0.5% of women [7, 8]. While this has been known for decades, colormaps that cannot be easily interpreted by those with CVD are still grossly over utilized [9, 10], perhaps because they are often the default colormaps in data processing software. This can cause misinterpretation for certain types of data, even for those with normal color vision (Fig 1) [11]. One possible solution for representing data is to simply switch to grayscale colormaps. Grayscale maps avoid the issues associated with color perception and can provide linear luminance with relation to underlying values. However, grayscale suffers from some viewing condition adaptations and has a small dynamic range due to the lower discernibility of the shades of gray to the human visual system. Humans without CVD can distinguish around ten million different colors versus only about thirty shades of gray [12]. CVD limits this range of colors dramatically (Fig 2), but even limited color vision can make use of the higher dimensionality of colorspace available when using non-grayscale colormaps when they are optimized correctly. This makes color an indispensable component to colormaps, as it enables the ability to see subtle changes in the underlying data (i.e. increases our visual perception precision due to the larger dynamic range).

Another important consideration for how we view color, with or without CVD, is how we perceive color differences. To represent a color, red-green-blue (RGB) values are most often used, but changes in these values are not linearly proportional to how we perceive color change. The sRGB colorspace was designed for employing RGB values in common digital displays [13] and is the standard space most often used to pass color information (as RGB values), even though it covers a limited portion of colors visible to a human with normal trichromatic (no CVD) vision [14, 15]. Because of its wide applicability and the widespread use of digital displays for viewing scientific data, this is the colorspace we both start from and return to (Fig 3, discussed in Materials and

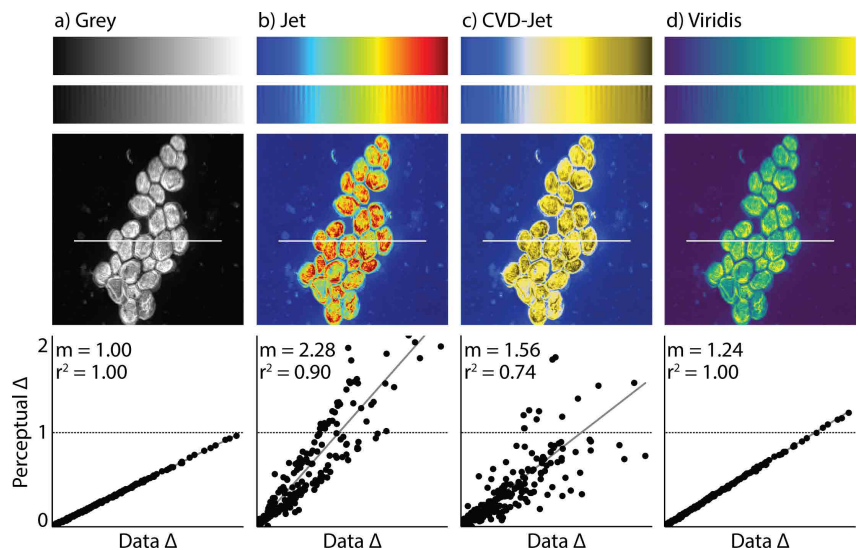

**Fig 1. Example of a misleading colormap.** Comparison between different colormaps overlaid onto the test image by Kovesi and a nanoscale secondary ion mass spectrometry image. Colormaps are as follows: (a) perceptually uniform grayscale, (b) jet, (c) jet as it appears to someone with red-green colorblindness, and (d) viridis [1], the current gold standard colormap. Below each NanoSIMS image is a corresponding "colormap-data perceptual sensitivity" (CDPS) plot, which compares perceptual differences of the colormap to actual, underlying data differences. $m$ is the slope of the fitted line and $r^2$ is the coefficient of determination calculated using a simple linear regression. An example of how the data may be misinterpreted are evident in the bright yellow spots in (b) and (c), which appear to represent significantly higher values than the surrounding regions. However, in fact, the dark red (in b) and dark yellow (in c) actually represent the highest values. For someone who is red-green colorblind, this is made even more difficult to interpret due to the broad, bright band in the center of the colormap with values that are difficult to distinguish.

https://doi.org/10.1371/journal.pone.0199239.g001

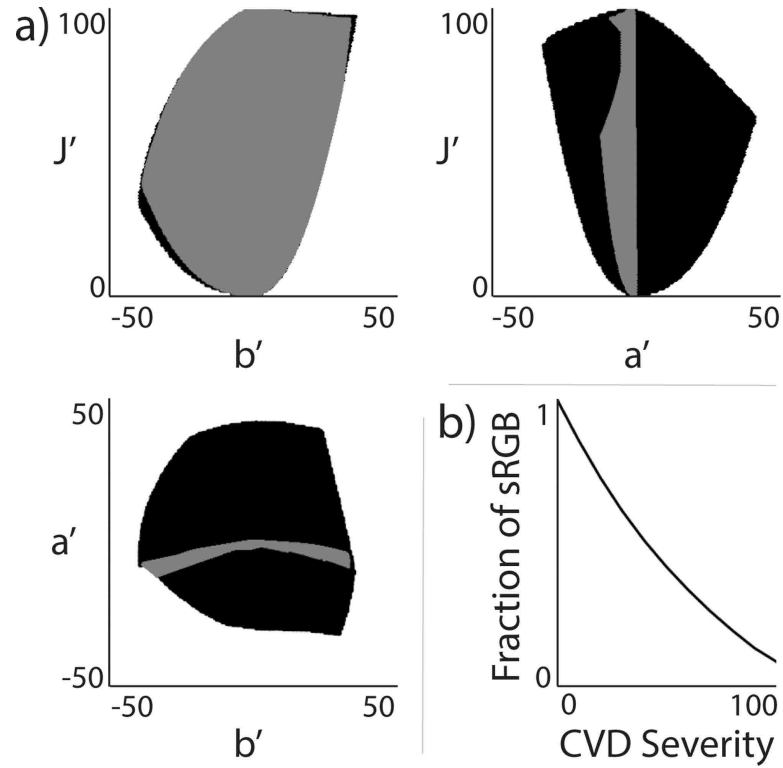

**Fig 2. CVD-safe colorspace in CIECAM02-UCS.** Visual of how limited color vision is for those with CVD. (a) 2D-representation of area of colorspace accessible to those without (black) and with (gray) complete red-green colorblindness as a function of the CIECAM02-UCS parameters ($J'$, $a'$, and $b'$). (b) Fraction of sRGB colors visible as a function of deuteranomaly severity. A severity of 0 corresponds to normal color vision whereas a severity of 100 corresponds to complete dichromacy (i.e. red-green colorblindness in this case).

https://doi.org/10.1371/journal.pone.0199239.g002

Methods); however, sRGB is not a suitable colorspace for colormap modification since understanding how humans truly perceive difference in color is essential in colormap design.

One of the first color spaces created to represent actual human perception of color was the CIE 1931 XYZ colorspace created by the International Commission on Illumination (CIE) [16]. This colorspace is still used today and well accepted; however, it does not account for variable viewing conditions. This led to the need for color appearance models (CAMs) which can take viewing conditions into account and use them to create new colorspaces. Phenomena dealt with using CAMs include chromatic and spatial adaptation and changes in the perception of a color due to factors such as hue and contrast changing based on their surroundings. An image appearance model, named iCAM, has also been created as a way to simulate more complex viewing conditions that cannot be achieved with traditional CAMs alone [17].

The comprehensive color appearance model CIECAM02 was created to better account for how we perceive color and is effectively the international gold standard [15, 18]. Luo et al. used this to create a new colorspace, named CIECAM02-UCS (Uniform Color Space), optimized to account for large and small differences in color [19]. In this colorspace, three values are used to describe a color: lightness ($J'$), a red-green correlate ($a'$), and a yellow-blue correlate ($b'$) (Fig 2). Equal Euclidean distances between values in this space leads to an equal color difference perception, enabling the straightforward creation of perceptually uniform colormaps;ºperceptually uniformº in that a difference in color space is intended to closely correspond monotonically to human perception of color difference. By converting between sRGB

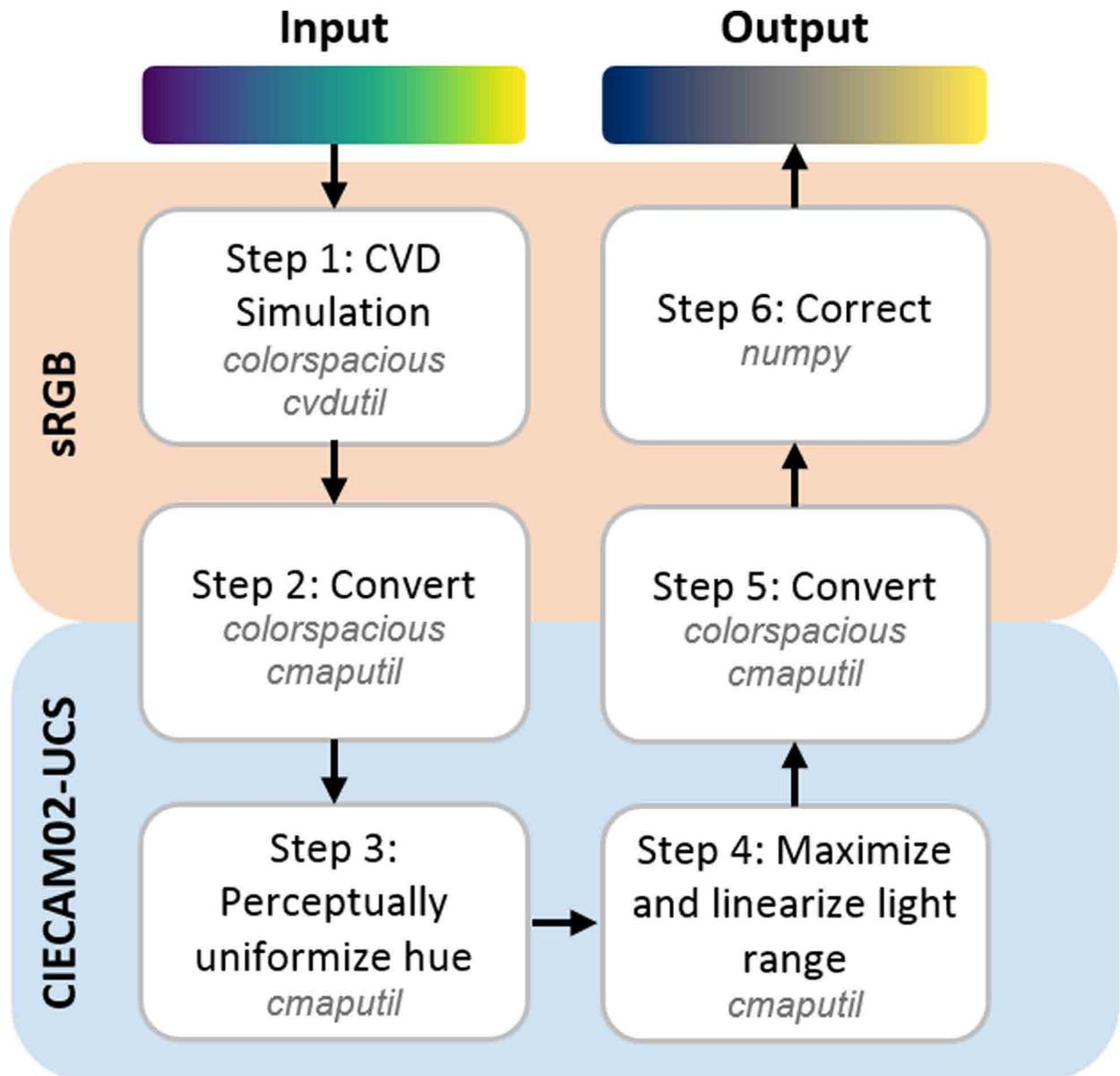

**Fig 3. Script pipeline.** Schematic of our script and how it optimizes colormaps for CVD. The colorspace, either sRGB or CIECAM02-UCS, where each operation takes place is shown along with the Python packages specifically required for each step.

https://doi.org/10.1371/journal.pone.0199239.g003

colorspace and CIECAM02-UCS colorspace, we can ensure colormaps are optimized for viewing via modern monitors, which cover sRGB space. This is an important consideration because reading science articles and viewing science data using digital displays is now ubiquitous.

To quantify the perceptual sensitivity of a colormap, we introduce the "colormap-data perceptual sensitivity" (CDPS) plot which is designed to compare perceptual differences to actual, underlying data differences (Fig 1). Specifically, the "Data Δ" is calculated by taking the absolute value of the difference between underlying image data points (in this case, the data points used were along the white line shown in each overlaid image). It is important to note this data has been normalized, but the same normalized image was used for the generation of each overlaid image and CDPS plot. The "Perceptual Δ" is the perceptual difference between colored pixels (calculated as the Euclidean distance between their points in CIECAM02-UCS space) of the overlaid image, normalized by dividing this distance by the slope of the grayscale case. How the overall slope in the CDPS plot compares to greyscale (which always has a slope of 1) indicates how far apart colors within the colormap are. Higher slopes correspond to colormaps that cycle through more colors, and therefore have a higher sensitivity between colors in the map. The coefficient of determination, $r^2$, calculated using a simple linear regression, indicates how well the changes represent the true underlying data differences. High $r^2$ show a better correlation between differences in the underlying data and the perceived color differences (i.e., colormaps with $r^2 = 1$ are perceptually uniform).

Colormaps have been used for almost 150 years [20] and design techniques have steadily evolved throughout this time due to increasing awareness of the effect colormaps can have on data interpretation [4, 5]. For an in-depth review on commonly accepted colormap design methods, please refer to the review by Dr. Peter Kovesi [21], which includes descriptions of different colormap types, the history of colorspace design, best practices for their design, and examples for their use. It is important to note that for this study, we focus on colormaps applicable for typical scalar scientific data, which assumes a need for monotonically increasing colormaps. This is not the case for all data. Non-monotonically increasing colormaps are well accepted for specific types of data, such as using a diverging colormap [21] to represent the correlation between a subject and its reference.

From Kovesi's review and others [1, 2, 22], there have been a few design principles for creating modern monotonically increasing colormaps. First, a linear increase in lightness should be used to avoid the perception of gradients that are not present. For example, colormaps can quickly change in brightness and hue, causing small changes in the underlying data to seem significant or, in regions where the values change too slowly, large changes to seem negligible. Along with avoiding this banding effect, a linear increase in brightness makes it straightforward to interpret which values are more significant than others and allows direct comparison between values within and (assuming the same scale is used) between images. Second, colors in the colormap should be equidistance from each other in colorspace, ensuring colors perceptually change at a constant rate (i.e., perceptually uniform). Being uniform in this aspect is a quantitative way of preventing regions of the colormap from changing too quickly or too slowly, which helps truly significant changes in data to be visually apparent. Colormaps created using these design principles can be evaluated by using the test image mentioned in Kovesi's review. This test image works by using a sine wave to iterate through the colors of the colormap, with increasing amplitude toward the top edge of the test image. Areas where the sine wave cannot be easily seen show the presence of regions where the color gradient is too small to properly distinguish the different color values. Additionally, inconsistent presence, or lack thereof, of the sine wave at each amplitude shows varying color gradients throughout the map. Our module, *cmaputil*, includes a function to create this test image as well as overlay a colormap on it.

While these design principles have led to the development of several wonderful, and increasingly popular, perceptually uniform colormaps (e.g., viridis in matplotlib [1], parula in MATLAB [23], and the cmocean package for oceanography applications [24]), we believe two additional critical design principles should be added. First, J′ should cover as large a range as possible within the bounds dictated by a′ and b′ (and, in the case for monitor-viewable colormaps, without leaving sRGB space), in order to increase perceptual distances between points and further improve the correlation of brightness with higher values. Second, all designs principles should be completed considering CVD.

We created a Python module called *cmaputil* that can automatically change a pre-defined colormap so that it complies with these recommendations considering a chosen CVD type (deuteranomaly, protanomaly, or tritanomaly). Note that most of those with CVD still have trichromatic vision but have a limited perception of colors due to one cone type having an altered range of sensitivity, causing it to overlap with another cone type, limiting the ability to distinguish colors detected by those two photoreceptors. Complete colorblindness, the most severe form of each type of CVD, is a complete absence of a cone. To our knowledge, our study here is the first to mathematically optimize a colormap specifically for viewing by those both with and without CVD. Below, we discuss the creation of our module and the many considerations that came along with its creation. We also present cividis and explain why we feel it is optimal for the typical scientific data set.

## Materials and methods

### Color vision deficiency simulation

Several CVD simulators are readily available within Python. Among these, we chose *colorspacious* which, to our knowledge, was the only module that also had the ability to convert between sRGB colorspace and CIECAM02 color spaces [1, 25]. The CVD simulation model used in *colorspacious* is from Machado et al. [26] and requires as input the CVD type as well as severity, both of which are easily controlled through this module. Specifically, CIECAM02-UCS was used for the colorspace for altering the colormap. Deuteranomaly was chosen as the CVD type because deuteranomaly, also referred to as a red-green color deficiency (protanomaly being another form of red-green color deficiency), is by far the most common form of CVD [7, 8]. Severity can also be controlled by adjusting it along a scale of 0±100, with 0 representing no colorblindness and 100 representing complete dichromacy. We chose to use a severity of 100 to ensure that anyone with either partial or complete red-green color deficiency could benefit from the same optimized colormap. The colorspace, CVD type, and CVD severity settings can all easily be changed within the code.

It is important to note other algorithms exist for simulating CVD and colorblindness [27, 28], including the "corresponding pair algorithm," which can account for different viewing conditions [29, 30]. We found the model [26] used here is the best for our purposes since it is not limited to dichromatic vision simulation. Different design considerations of this model that make it useful across many applications are also discussed in the original paper [26].

### cmaputil: Python-based colormap utilities

The scripts referred to in this paper can be found at https://github.com/pnnl/cmaputil or can be downloaded using PyPI (`pip install cmaputil`). Our module has other capabilities beyond those discussed here, some of which are described in a previous paper [3]. The process used for testing and optimizing colormaps is provided as an example script with the code.

Python (v 2.7.10), with the packages OpenCV2 (v 3.1.0, opencv.org), numpy (v 1.9.3) [31], matplotlib (v 1.5.0) [32], and *colorspacious* (v 1.1.0) [25], was implemented using WinPython (v 2.7.10, winpython.github.io). IPython (v3.2.0) [33], an enhanced Python shell, was used within the Scientific Python Development Environment (Spyder, v2.3.5.2, github.com/spyder-ide) for interactively analyzing data and creating figures.

An overview of our pipeline is shown in Fig 3. A visual example iteration of a colormap as it is being adjusted is shown in Fig 4. For the implementation of our pipeline, a colormap (represented as an array of RGB values with each column representing a new color) is required as input. There are many colormaps readily available for download or via built-in modules such as matplotlib. Colormaps can also be created using custom techniques or the graphical user interface created by Dr. Smith's lab, *viscm* [1]. One useful part of *viscm* is that it shows where colors lie within CIECAM02-UCS color space.

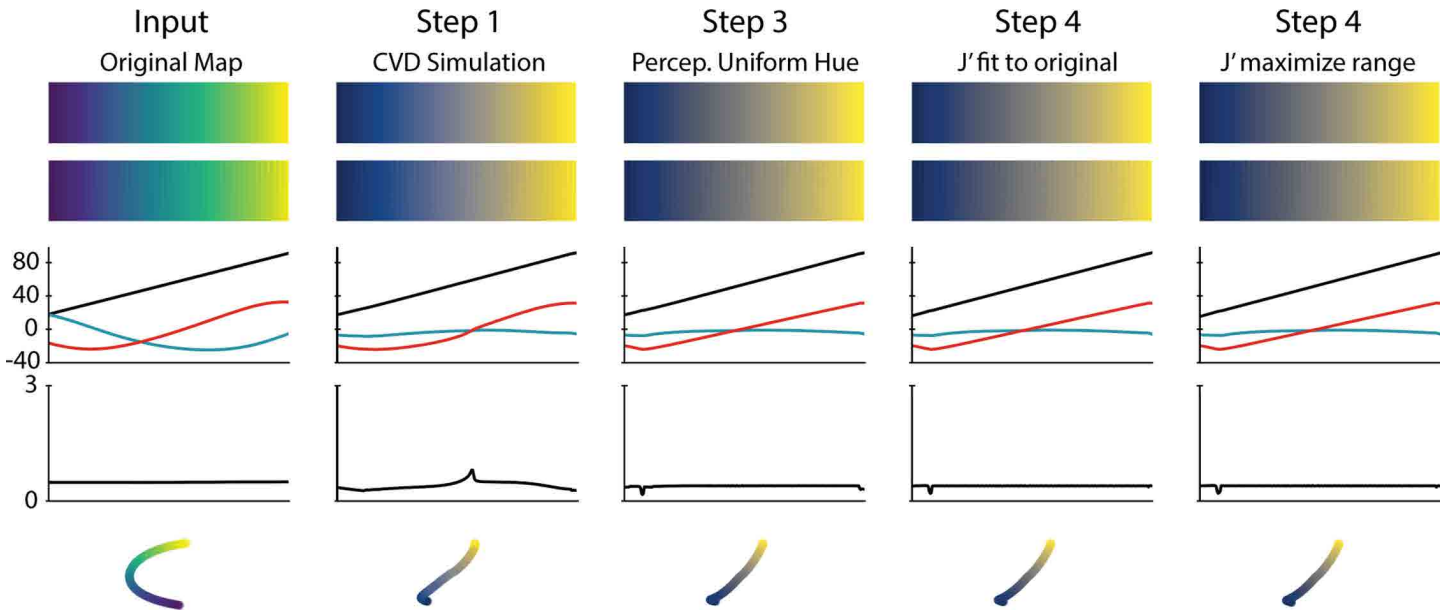

**Fig 4. Colormap adjustment iterations.** In this example, the viridis colormap is taken through each stage of our pipeline. From top to bottom, the image plotted is the colormap (i) as it was input, (ii) overlaid on the test image discussed by Peter Kovesi [21], and (iii-v) based on the method presented by the Smith group [1], these show the values of this colormap in CIECAM02-UCS space, with (iii) comparing individual values J′ (black), a′ (blue), and b′ (red) across the map, (iv) showing the perceptual deltas between each point on the map, calculated as the Euclidean distance between each point, and (v) providing a three dimensional view of the colormap in this space.

https://doi.org/10.1371/journal.pone.0199239.g004

The input colormap is converted to the RGB values as seen by someone with CVD, based on the CVD type defined in the module. From here, these values are converted to CIECAM02-UCS and the colormap is optimized by (i) interpolating a′ vs. b′ to generate points equidistance from each other, then (ii) linearizing and maximizing the J′ range. Once this is complete, the colormap is brought back to sRGB colorspace for easy sharing and application within other modules. Details for the optimization follow.

The a′ and b′ arrays are interpolated using 10,000 points each. The total arc length of a′ vs. b′ is then calculated. This value is divided by 255 to calculate the desired distance to be achieved between each set of points within the final colormap to ensure perceptual uniformity. The number of points in common colormaps is 256, but the interpolation function can easily be changed for generating 512, 1024, or other sized colormaps. Once this distance is found, all 10,000 sets of (a′, b′) points are iterated through to select points that are this distance apart. It is important to note this method is used to ensure the total arc length through a′b′ space is the same before and after this interpolation function so the original hues are used and the colormap still travels through the same path in CIECAM02-UCS colorspace.

J′ is then linearized using two independent methods. The first method simply fits a line to the original J′ trend using linear regression. For the second method, each (a′, b′) point is iterated through to find the minimum and maximum J′ values that can be matched with that point to create a valid RGB value (i.e., a value between 0 and 1). Once these ranges are found for all 256 (a′, b′) points, a line is fitted within these bounds with the steepest slope possible.

Please refer to S1 File to see iterative changes made to several example colormaps.

### Mass spectrometry imaging data

The yeast images in Figs 2 and 5 were collected using mass spectrometry imaging of Baker's Yeast (Red Star Yeast), drop cast (DI water) onto a silicon wafer and dried, performed with high-lateral resolution secondary ion mass spectrometer (NanoSIMS, Cameca NanoSIMS 50L, Gennevilliers Cedex, France), which is housed in the Environmental Molecular Science Laboratory. Sample preparation and analysis was performed similarly to Renslow et al. [34]. Briefly, prior to analysis the sample was coated with 10 nm of Au to minimize charging during analysis [3]. High current sputtering was performed with the $Cs^+$ primary ion beam prior to collecting data, where samples were dosed with ~2 x $10^{16}$ ions/$cm^2$ to achieve sputtering equilibrium [3]. A ~1.5 pA $Cs^+$ primary ion beam was used for all analysis, and the $^{12}C^{12}C^-$, $^{12}C^{13}C^-$, $^{12}C^{14}N^-$, and $^{12}C^{15}N^-$, and $^{31}P^-$ secondary ions were detected simultaneously. The data visualized in this manuscript is the $^{12}C^{14}N^-$ ion count data. The imaging area was 40 μm x 40 μm, acquired at 256 pixels x 256 pixels, with 2 ms/pixel over nine planes.

### COMSOL Multiphysics® modeling data

COMSOL Multiphysics®, (COMSOL, Inc., Burlington, MA, USA), a finite element analysis software package, is used throughout many scientific research areas [35–38]. Our optimized colormap, cividis, will be available in COMSOL v5.3a and higher. Here, we demonstrate the use of cividis for displaying a velocity map in a simple fluid flow model (Fig 5). The model, available in the Fluid Dynamics examples provided with the software (and available at https://www.comsol.com/model/flow-past-a-cylinder-97, which includes the *cylinder_flow.mph* files and details in *models.mph.cylinder_flow.pdf*), simulates the time-dependent flow past a cylinder. The model examines unsteady, incompressible flow past a long cylinder placed in a

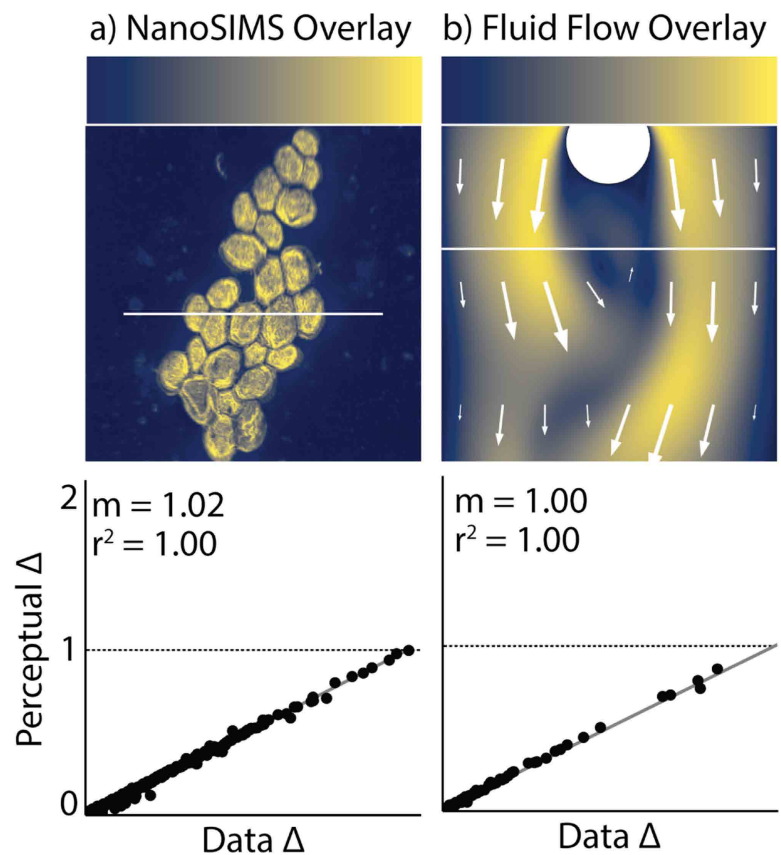

**Fig 5. Our optimal colormap, cividis.** Colormap shown overlaid onto a) NanoSIMS image and b) fluid velocity map from COMSOL. Below is each corresponding CDPS plot for data along the white lines.

https://doi.org/10.1371/journal.pone.0199239.g005

channel at right angle to the oncoming fluid. The cylinder is offset somewhat from the center of the flow to make the steady-state symmetrical flow unstable. In this simulation the Reynolds number equals 100, which gives a developed Karman vortex street; but the flow is still not fully turbulent.

## Results and discussion

### Considerations for implementation

Linearizing J′ was more complex than initially predicted due to values in CIECAM02-UCS not always mapping to sRGB colorspace. Therefore, we introduced the ability to linearize J′ in two different ways: (i) fit to the original J′ line as closely as possible without considering how these values will map back to RGB colorspace or (ii) maximize the range of J′ based on valid (J′, a′, b′) to (R, G, B) mappings. The first option ensures the colormap returned is as close as possible to the original whereas the second option allows the ends of the colormap to be as far apart as possible which can lead to larger perceptual differences between colors and an increased ability to precisely distinguish between them. It is important to note the first method can fail due to invalid mappings and the second method can fail due to the inability to fit a straight line through the available J′ range. We found maximizing the J' range worked best in most cases. However, the user will need to use the method that is most appropriate for their specific use.

There are two unavoidable issues in *cmaputil* of which users should be made aware. One issue, which arises during conversion from CIECAM02-UCS to sRGB colorspace, is that sets of J′a′b′ values can map to invalid RGB values, as noted above. This is due to the fact that CIECAM02-UCS covers much more than sRGB space since it does not encompass all human visible colors [14, 15]. When the mapping is invalid, no errors are returned by *colorspacious* during conversion. To avoid invalid RGB values, each time a color array comes back from CIECAM02-UCS, we use an absolute colorimetric rendering intent [39][40] for gamut mapping by ensuring all values are set to be between 0 and 1 (note we are using the 0±1 scale rather than 0±255 to represent the full span of RGB colorspace). This means color values between 0 and 1, which are valid, are not changed. If outside this range, they are then replaced with a 0 or 1, depending on which value is closer. Absolute colorimetic rendering intent can lead to large departures from the intended color, especially when the invalid color values are far from 0 or 1. This is an additional reason why we provide two methods of linearizing J′. From our experience, if one linearization method fails, the other method can be chosen by the user.

The second issue is that the perceptual deltas are not always perfectly linear even though they are optimized to be constant across the colormap. This is due to the fact *cmaputil* manipulates colors in CIECAM02-UCS then converts back to sRGB to determine the RGB values of the new colormap. To analyze how the colormap travels through CIECAM02-UCS (such as in Fig 4), the RGB values of the newly optimized colormap are converted back to CIECAM02-UCS space, rather than using the original CIECAM02-UCS values available prior to conversion to sRGB. This may seem counterintuitive since CIECAM02-UCS values are converted to sRGB and back before determining their perceptual deltas but this is important since they may initially change upon the first conversion to RGB if they fall outside sRGB colorspace. For example, in Fig 4, the colormap created in Step 4 is supposed to have a perfectly linear perceptual delta across its length, however, there is a slight bump around the 20$^{th}$ value in the colormap. In this case, this is because the value for R in sRGB colorspace upon conversion of CIECAM02-UCS to sRGB is about -0.038. Following an absolute colorimetric rendering intent, this value gets set to zero, leading to the slight difference (0.47% average error, nearly negligible) between the original and final J′a′b′ value.

Even though the perceptual deltas are not perfectly flat, we felt our goal to provide color values that are perceptually uniform is still accomplished as any imperfections in this line are relatively small (<0.2, as compared to perceptual deltas of >0.8 seen in other colormaps such as jet).

**An optimized colormap**

We used *cmaputil* to create several example colormaps. We identified one colormap in particular to be optimal for viewing by those with or without CVD, which we name cividis (Figs 4 and 5), generated by optimizing the viridis colormap and selecting the J' linearization that maximizes the range of J'. We chose this map due to its wide range of colors, resulting from a wide range of J' values while still changing b' significantly, and overall sharpness when overlaid onto complex images. Other optimized colormaps produced using this method had a more limited J' range due to spanning more of the b' space. While spanning more hues is preferable as well, we felt this colormap had the best balance of hue and lightness range. For the RGB color values of this colormap, please refer to S2 File. This colormap is also close to optimal for protanomaly and tritanomaly, as shown in S4 File, where figures are plotted with each of the forms of CVD with severity set to 100.

Right now, viridis is seen as the gold standard since it follows each of the design principles discussed above; however, it is optimal for those with normal vision and not CVD. While its CVD-simulated counterpart is close to optimal, *cmaputil* helped to further improve upon this by forcing the perceptual delta to be as close to flat as possible and by increasing the light range covered. At this time, it is preferable to have both viridis and cividis available to the scientific community as viridis is more aesthetically pleasing to some and covers more colors for those with normal vision, allowing greater visual perception sensitivity (Fig 1 viridis CDPS plot vs. the Fig 5 CDPS plot).

While it may take some time for the full scientific community to both be aware of the need to choose appropriate colormaps and agree on preferred colormaps, we hope the code we provide here can help with this transition by allowing others to experiment with the different aspects of colormap design and see how the various characteristics of a colormap affect its interpretation. Our code can also be leveraged for creating non-monotonically increasing colormaps, by adjusting the J' linearization function, introducing the ability to create colormaps for datasets where this is the optimal choice, such as when a diverging colormap is preferred.

Our team has been working to ensure this colormap is easily accessible. As a result, COMSOL [41], the OpenMIMS plugin for ImageJ [42], and Fiji [43], will be adding cividis as an option among their colormaps. For those who already have a copy of ImageJ, the LUT version of cividis is provided as S3 File. We will also be reaching out to other software teams to further increase awareness about our colormap and important considerations in colormap design. Ideally, we hope to reduce the number of non-perceptually uniform and rainbow-style colormaps being used as default in scientific data analysis software, and to increase the availability of CVD-friendly colormaps so the estimated more than 600 million individuals worldwide with CVD can interpret and perceive data like the rest of the population.

An important consideration about our colormap moving forward is it may become obsolete due to monitors and screens being designed to display larger color gamuts, i.e. colors outside sRGB colorspace [44, 45]. Our module can again be used for this purpose simply by changing the sRGB colorspace setting to the more modern colorspace viewed on future screens (e.g., 8K Ultra HD screens, which may eventually cover nearly twice as much natural color space as the current sRGB). Additionally, this could better enable *cmaputil* to create perfectly straight perceptual delta lines, even after conversion to and from CIECAM02-UCS, due to more colors being available than in sRGB color space.

### Future work

A downside of cividis, as reported by colleagues, is its minimal coverage of different colors: varying straight from blue to yellow rather than cycling through other colors, as viridis does. This keeps cividis from being as aesthetically pleasing as viridis. Of course, this is because those who have a form of CVD cannot see these colors the way those with normal vision can. However, since normal color vision is more common, using more colors is often desired for representation of data and for increasing visual perception precision through use of a larger dynamic color range. An area of research we are pursuing is the ability to cycle through more colors while still keeping *both* normal color vision and deuteranomaly (ranging from a mild form to complete dichromacy) perceptions of the colormap optimal. Furthermore, by using CDPS plots (e.g., shown in Figs 1 and 5) to judge our success, we plan to maximize the slope ($m$, thus maximizing perceptual sensitivity) while keeping the colormap perceptually uniform (i.e. $r^2 \approx 1$). Viridis is already close to accomplishing this since the perceptual delta of the colormap, as perceived by someone with complete red-green colorblindness, has only a small imperfection toward the center. Through the next version of *cmaputil* we hope to create a ªviridis 2.0º, and other aesthetically pleasing colormaps, that are perceptually linear for all people. It is not clear yet if this task is possible, as those with deuteranomaly but still have trichromatic vision will be able to see some of the additional colors to varying extents, making it difficult to optimize for all possible severities. We will try different variations of colormap creation and optimization to find if this can be done.

## Supporting information

**S1 File. Colormap results.** Zip file of example optimization figures for multiple colormaps we tested. This includes the iteration image (colormap conversion to CVD, interpolation of a′ vs. b′, and the two J′ linearization methods), a′ vs. b′ change overlay, and a comparison of the two fits used for J′.
(ZIP)

**S2 File. Cividis.** Table with all 256 colormap values for our optimal colormap, cividis.
(TXT)

**S3 File. LUT Cividis.** Table with all 256 colormap values for our optimal colormap, cividis. This format matches the LUT format required by the ImageJ software. To add this to your copy of ImageJ, simply paste it in ImageJ/luts. Note this will be included in upcoming Fiji versions.
(LUT)

**S4 File. Cividis overlay examples.** Three images overlaid with the cividis colormap as it would appear with one of the three forms of CVD (deuteranomaly, protanomaly, and tritanomaly), severity 100. Middle image,"4x autofluor.tif", collected from imagej.nih.gov/ij/docs/examples/IJ-M&M08-Figures.zip.
(PNG)

## Acknowledgments

We would like to thank our colleagues who expressed the need for this project, as well as provided advice and feedback. This research was partially supported by the Genomic Science Program (GSP), Office of Biological and Environmental Research (OBER), the U.S. Department of Energy (DOE), and is a contribution of the Pacific Northwest National Laboratory (PNNL) Foundational Scientific Focus Area (SFA). A portion of this research was performed in the W.

## Author Contributions

**Conceptualization:** Ryan S. Renslow.

**Data curation:** Christopher R. Anderton.

**Formal analysis:** Ryan S. Renslow.

**Funding acquisition:** Jamie R. Nuñez, Ryan S. Renslow.

**Investigation:** Jamie R. Nuñez, Ryan S. Renslow.

**Methodology:** Jamie R. Nuñez, Christopher R. Anderton.

**Resources:** Christopher R. Anderton.

**Software:** Jamie R. Nuñez, Ryan S. Renslow.

**Supervision:** Ryan S. Renslow.

**Validation:** Jamie R. Nuñez.

**Visualization:** Jamie R. Nuñez, Ryan S. Renslow.

**Writing ± original draft:** Jamie R. Nuñez, Ryan S. Renslow.

**Writing ± review & editing:** Jamie R. Nuñez, Christopher R. Anderton, Ryan S. Renslow.